%% file: main.tex
\documentclass{article}

\usepackage[preprint]{neurips_2024}

\usepackage[utf8]{inputenc} 
\usepackage[T1]{fontenc}    
\usepackage{hyperref}       
\usepackage{url}            
\usepackage{booktabs}       
\usepackage{amsfonts}       
\usepackage{nicefrac}       
\usepackage{microtype}      
\usepackage{xcolor}         
\usepackage{amsmath}
\usepackage{float}
\usepackage{tabularray}
\usepackage{multirow}
\usepackage{booktabs, caption, array}
\usepackage{amssymb}
\usepackage{lipsum}         
\usepackage{CJKutf8}        
\usepackage{nomencl}
\usepackage{glossaries}
\usepackage{xspace}
\usepackage{multirow}
\usepackage{graphicx}
\usepackage{bbding}
\usepackage{pifont}
\usepackage{algorithm}
\usepackage{subfigure}
\usepackage{enumitem}
\usepackage{multicol}
\usepackage{caption}
\usepackage{array}
\usepackage{fp}
\usepackage{multirow}   
\usepackage{makecell}
\usepackage{flushend}
\usepackage{cases}
\usepackage{color}
\usepackage{algpseudocode}
\usepackage{listings}
\usepackage{mathrsfs}
\usepackage{longtable}
\usepackage{colortbl}
\usepackage{bm}
\usepackage{caption}
\setlist[itemize]{noitemsep, topsep=0pt}
\usepackage{arydshln}
\usepackage{lipsum}
\definecolor{codegreen}{rgb}{0,0.3,0.6}
\definecolor{codegray}{rgb}{0.5,0.5,0.5}

\newcommand{\ie}{\emph{i.e.,}\xspace}

\newcommand{\aka}{\emph{a.k.a.,}\xspace}

\newcommand{\paratitle}[1]{\vspace{1.5ex}\noindent\textbf{#1}}

\newcommand{\ignore}[1]{}

\usepackage{tcolorbox}
\definecolor{darkorange}{RGB}{255, 140, 0}
\definecolor{lightgreen}{RGB}{145, 204, 117}
\definecolor{lightyellow}{RGB}{250, 200, 88}
\definecolor{lightred}{RGB}{238, 102, 102}
\definecolor{lightblue}{RGB}{115, 192, 222}
\newtcolorbox{promptbox}[3][Judge Prompt]{
colback=black!5!white,
arc=5pt, 
boxrule=0.5pt,
fonttitle=\bfseries,
title=#1, 
before upper={\small}, fontupper=\fontfamily{ptm}\selectfont,
colframe=#2,
label=#3,
}

\title{R1-Searcher:  Incentivizing the Search Capability in LLMs via Reinforcement Learning}

\author{%
   Huatong Song$^{1}$\thanks{Equal contribution.}~,
  Jinhao Jiang$^{1*}$,
  Yingqian Min$^{1}$,
  Jie Chen$^{1}$,
  Zhipeng Chen$^{1}$,\\
  \textbf{Wayne Xin Zhao$^{1}$\thanks{Correspondence to Wayne Xin Zhao.}~, 
  Lei Fang$^2$,} 
   \textbf{Ji-Rong Wen}$^{1}$
  \\
  $^1$Gaoling School of Artificial Intelligence, Renmin University of China.\\
  $^2$DataCanvas Alaya NeW\\
  \texttt{\{songhuatong123, jiangjinhao\}@ruc.edu.cn}\\
  \texttt{batmanfly@gmail.com}
}

\begin{document}
\maketitle
\begin{abstract}
Existing Large Reasoning Models (LRMs) have shown the potential of reinforcement learning (RL) to enhance the complex reasoning capabilities of Large Language Models~(LLMs). While they achieve remarkable performance on challenging tasks such as mathematics and coding, they often rely on their internal knowledge to solve problems, which can be inadequate for time-sensitive or knowledge-intensive questions, leading to inaccuracies and hallucinations. 
To address this, we propose \textbf{R1-Searcher}, a novel two-stage outcome-based RL approach designed to enhance the search capabilities of LLMs. This method allows LLMs to autonomously invoke external search systems to access additional knowledge during the reasoning process. Our framework relies exclusively on RL, without requiring process rewards or distillation for a cold start.
Our experiments demonstrate that our method significantly outperforms previous strong RAG methods, even when compared to the closed-source GPT-4o-mini.
The code is available at \url{https://github.com/RUCAIBox/R1-Searcher}.

\begin{figure}[ht]
\centering
    \centering
    \includegraphics[width=0.85\linewidth]{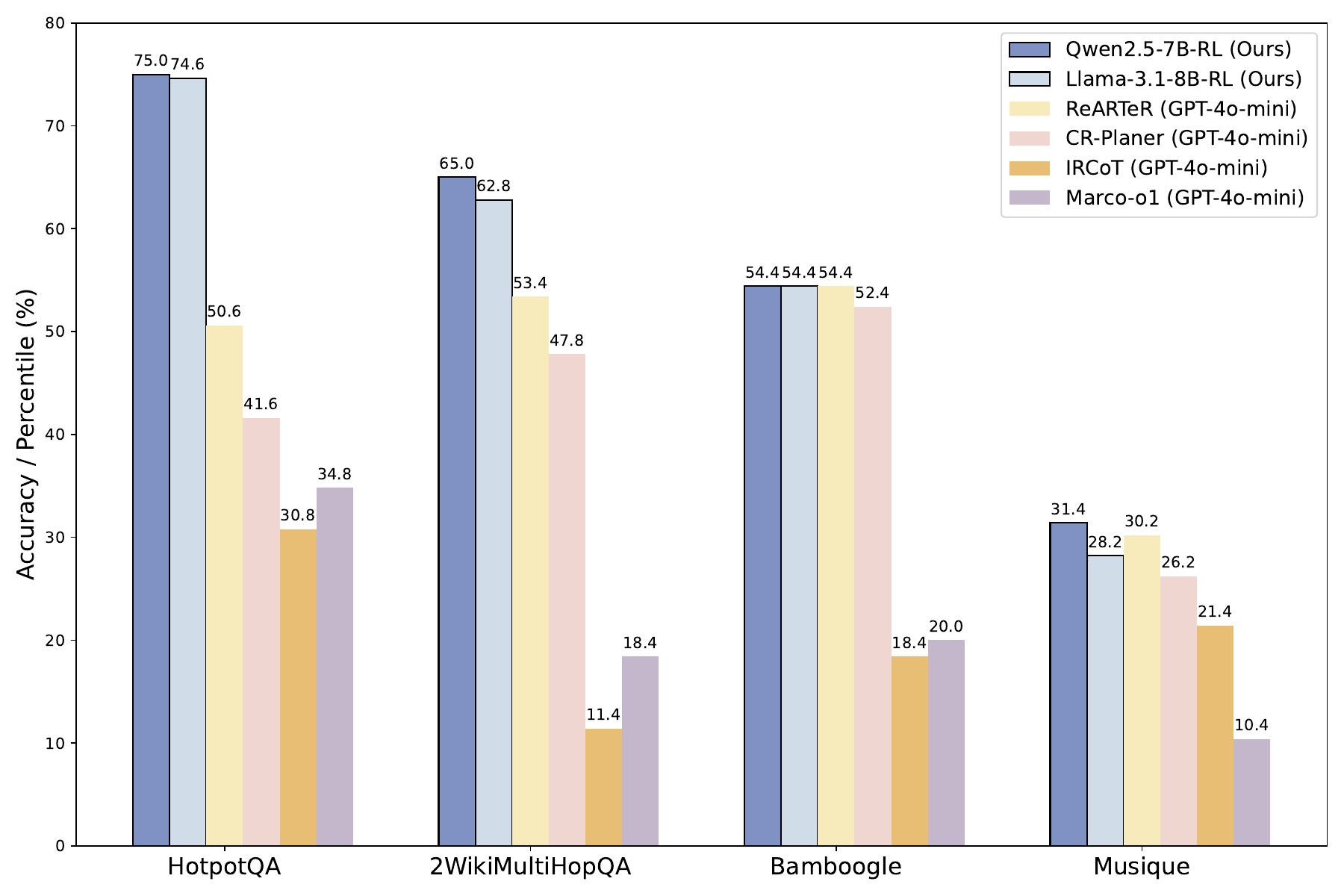}
\caption{Performance comparisons between R1-Searcher and other methods on four multi-hop QA benchmarks. R1-Searcher achieves significant performance improvement on each dataset. The metric is LLM-as-Judge Score.}
    \label{fig:benchmarks}
\end{figure}
\end{abstract}

\input{sections/introduction}
\input{sections/method}
\input{sections/exp}

\input{sections/research_question}

\input{sections/case_study}

\input{sections/conclusion}

\input{sections/future_work}


\bibliographystyle{unsrt}
\bibliography{ref.bib}

\end{document}

%% file: sections/introduction.tex
\section{Introduction}
\label{sec-intro}

Large Reasoning Models (LRMs), such as OpenAI-o1~\cite{openaio1}, Deepseek-R1~\cite{deepseekr1} and Kimi-k1.5~\cite{kimik1.5}, have demonstrated the significant impact of reinforcement learning (RL) in enhancing the reasoning capabilities of large language models (LLMs)~\cite{still-3}. However, since they primarily rely on their internal knowledge, these models may struggle with open-ended tasks, particularly those involving knowledge-intensive questions~\cite{yang2018hotpotqa, ho2020constructing}, private information in local databases~\cite{wang2025omnievalomnidirectionalautomaticrag, lee2024multirerankermaximizingperformanceretrievalaugmented}, and time-sensitive issues~\cite{press2023measuring, he2025mintqamultihopquestionanswering}. This reliance may easily lead to inaccuracies and hallucinations. Therefore, it is crucial to enable LLMs to access external information during the reasoning process to achieve more deliberative reasoning~\cite{rag-star}. 

To address this issue, extensive research has focused on augmenting LLMs with external information sources (\aka retrieval-augmented generation (RAG)~\cite{gao2024retrievalaugmentedgenerationlargelanguage, ragsurvey}). Early approaches emphasize specific prompting strategies to guide LLMs in iterative question decomposition, query generation, and sub-question answering~\cite{li2025searcho1agenticsearchenhancedlarge, self-rag, teng2025atomthoughtsmarkovllm}. While effective, these complex prompt designs may rely on closed-source LLMs for achieving optimal performance. Subsequent studies investigate to distill this capability into smaller LLMs through supervised fine-tuning (SFT)~\cite{corag}. However, recent findings suggest that SFT-based distillation can cause models to memorize solution paths, limiting their generalization to novel scenarios~\cite{chu2025sftmemorizesrlgeneralizes}. Recent proposals include a test-time scaling method~\cite{rag-star, li2024can}, notably employing the Monte Carlo Tree Search (MCTS) framework to enhance solution-finding by expanding the search space during inference. Despite its promise, this approach incurs significant inference overhead, reducing its practicality for widespread use. Therefore, we propose integrating an external retrieval environment during training, enabling models to explore and learn to effectively utilize retrieval for problem-solving. This approach aims to incentivize the search capability in LLMs, thereby enhancing LLMs' generalization and improving inference efficiency.


In this paper, we introduce \textbf{R1-Searcher}, a novel framework to enhance the RAG capabilities of LLMs with RL. Our core motivation is to incentivizing the search capability in LLMs via exploring with an external retrieval environment. To implement it, we design a two-stage, outcome-based RL approach, enabling the model to freely explore how to invoke an external retrieval system to acquire relevant knowledge during the reasoning process through a tailored reward design. Specifically, in the first stage, we employ the retrieve-reward to incentivize the model to conduct retrieval operations without considering the final answer accuracy. In this way, the LLMs can quickly learn the correctly retrieval invocation format. In the second stage, we further introduce the answer reward to encourage the model to learn to effectively utilize the external retrieval system to solve question correctly. Our method relies solely on outcome-based RL, allowing the model to learn autonomously through exploration and learning without requiring any distillation or cold start with SFT. To support the exploration between LLMs and the external retrieval environment during the training process, we further propose a modified RL training method based on Reinforce++~\cite{reinforce++} with RAG-based rollout and retrieval mask-based loss calculation.

We conduct extensive experiments to verify the effectiveness of our method using various LLM backbones on four representative benchmarks, based on Llama-3.1-8B-Instruct and Qwen-2.5-7B-Base. Notably, our method surpasses the strong baseline with GPT-4o-mini (\ie ReARTeR) by up to 48.22\% on HotpotQA and 21.72\% on 2Wiki when using Qwen-2.5-7B-Base. To access generalization capability, we evaluate our method on the Bamboogle dataset using an online search, which is not seen during training. Our model achieved an 11.4\% performance improvement on Bamboogle compared to the Search-o1~\cite{li2025searcho1agenticsearchenhancedlarge} with 32B parameters.

Our principal contributions are as follows:

$\bullet$ We introduce R1-Searcher, which utilizes a two-stage RL framework to enable autonomous retrieval during the reasoning process.

$\bullet$ Extensive experiments on four multi-hop datasets show that R1-Searcher consistently and significantly surpasses existing RAG methods, even close-sourced GPT-4o-mini.

$\bullet$ Our approach exclusively employs RL for training without any distillation or cold start, while showing great generalization to out-of-domain datasets and online search scenarios. It is effective for both base and instruction-tuned models.

%% file: sections/method.tex
\section{Method}
\label{method}

\subsection{Data Selection}

In this study, we aim to enhance the search capabilities of LLMs for problem-solving within a retrieval environment using an outcome-based RL method. However, the independence of the retrieval environment may lead to issues that exceed its query scope during the RL training process, posing challenges to successful problem resolution and affecting the training efficiency. To address this, we conduct data selection and incorporate questions with varying difficulty levels.

Specifically, we select training data from the training sets of two representative multi-hop question answering datasets, namely HotpotQA~\cite{yang2018hotpotqa} and 2WikiMultiHopQA~\cite{ho2020constructing}. We utilize the {Qwen-2.5-7B-Instruct}~\cite{qwen2025qwen25technicalreport} model, equipped with a local retrieval system, and prompt the model (Prompt~\ref{prompt:selection}) in solving questions from these datasets using the external retrieval system. The prompt is provided below. Based on the number of rollouts required to correctly answer a question, we categorize the data into three levels of difficulty: easy (fewer than 10 rollouts), medium (10 to 20 rollouts), and difficult (more than 20 rollouts). These difficulty levels are then combined as delineated in Table~\ref{tab:training_set} to construct our training dataset.


\begin{table*}[h]
\centering
\small
\begin{tabular}{lcccc}
\toprule
\textbf{Stage} & \textbf{Dataset} & \textbf{Easy} & \textbf{Medium} & \textbf{Difficult} \\
\midrule
\multirow{2}{*}{\textbf{Stage-1}} & HotpotQA & - & 200 & -  \\
 & 2WikiMultiHopQA & - & 150 & - \\
\multirow{2}{*}{\textbf{Stage-2}} & HotpotQA & - & 2561 & 2000  \\
 & 2WikiMultiHopQA & - & 1087 & 2500 \\
\bottomrule
\end{tabular}
\caption{The information of the data used during RL training.
}
\label{tab:training_set}
\end{table*}

\begin{promptbox}[System Prompt for Data Selection]{lightgreen}{prompt:selection}
You are a helpful assistant. Given a question, you should answer it by first thinking about the reasoning process in the mind and then providing the final answer. The output format of reasoning process and final answer are enclosed within <think> </think> and <answer> </answer> tags, respectively, i.e., "<think> reasoning process here </think><answer> final answer here </answer>". You should perform thinking with decomposing, reflecting, brainstorming, verifying, refining, and revising. Besides, you can perform searching for uncertain knowledge if necessary with the format of "<|begin\_of\_query|> search query (only keywords) here <|end\_of\_query|>"."""
Then, the search system will provide you with the retrieval information with the format of "<|begin\_of\_documents|> ...search results... <|end\_of\_documents|>".
\end{promptbox}

\subsection{Two-Stage Outcome-based Reinforcement Learning}

To progressively improve the search capabilities of LLMs, we propose a two-stage outcome-based RL training method. In Stage-1, the model is trained to effectively utilize an external retrieval system. In Stage-2, the model is trained to incorporate search during the reasoning process to accurately solve questions.

\subsubsection{Reward Design}\label{format_reward}

Due to the absence of intermediate annotations in the training data, the RL process is primarily influenced by outcome rewards. By assigning distinct rewards across two stages, the model progressively learns to invoke the external retrieval system and effectively integrate retrieved documents into the reasoning process to answer questions.

In Stage-1, the reward function comprises a retrieval reward and a format reward. The primary goal here is to enable the model to recognize its ability to invoke the external retrieval system and learn its utilization, without considering the correctness of the model's answers. The model is explicitly encouraged to initiate search queries, and thus, no answer reward is assigned at this stage.
Specifically, the retrieval reward is defined as follows:
\begin{equation}
    R_{retrieval} =
\begin{cases}
\text{0.5}, & n \geq 1 \\
\text{0}, & n = 0
\end{cases}
\end{equation}
where $n$ represents the number of retrieval invocations. For the format reward, we first define the correct format as follows:
\begin{enumerate} 
\item The model's thinking process and final answer should be enclosed within the \texttt{<think>...</think>} and \texttt{<answer>...</answer>} tags, respectively. Additionally, only the final short answer is permitted within the \texttt{<answer>...</answer>} tag.

    \item The generated output must be free of any garbled or unreadable content.

    \item When invoking retrieval, the model should propose a query and encapsulate the query within the \texttt{<begin\_of\_query>...</end\_of\_query>} tags. Furthermore, the model is unable to generate documents directly without invoking retrieval.
\end{enumerate} 




Based on the above format requirements, the format reward is defined as follows:
\begin{equation}
    R_{format} =
\begin{cases}
\text{0.5}, & \text{if the format is correct} \\
\text{0}, & \text{if the format is incorrect}
\end{cases}
\end{equation}

Therefore, the final reward of Stage-1 is the sum of the retrieval reward and format reward. 

In Stage-2, we eliminate the retrieval reward and incorporate the answer reward. We apply the same format judgment criteria as in Stage-1, but with different penalties:
\begin{equation}
R_{format}^{'} =
\begin{cases}
\text{0}, & \text{if the format is correct} \\
\text{-2}, & \text{if the format is incorrect}
\end{cases}
\end{equation}
For the answer reward, we utilize the F1 score of the ground-truth answer and predicted answer, which is calculated as follows:
\begin{equation}
R_{answer} =
\frac{2*IN}{PN+RN}
\end{equation}

where \textit{PN} represents the word count of the predicted answer, \textit{RN} denotes the word count of the reference answer, and \textit{IN} indicates the word count of the intersection between the two answers.

Therefore, the final reward of Stage-2 is the sum of the answer reward and the format reward.

\subsubsection{Training Algorithm}

Our training algorithm is based on the Reinforce++ algorithm, which we have modified to suit our retrieval-augmented generation scenario. During the reasoning process, the model engages an external retrieval system to solve problems, receiving a reward for correct solutions. We enhance the model's ability to utilize retrieval during the reasoning process by maximizing this reward. Our goal is to enable the model to autonomously access external knowledge when faced with uncertainty, effectively integrating reasoning and retrieval. To incorporate retrieved documents seamlessly and ensure rational model optimization, we implement two modifications to the original algorithm: \textit{RAG-based Rollout} and \textit{Retrieval Mask-based Loss Calculation}.

\paratitle{RAG-based Rollout.} As demonstrated in Prompt~\ref{prompt:system}, we guide the model to utilize the external retrieval system during the generation process by employing the tags \texttt{<begin\_of\_query>...<end\_of\_query>} to indicate the invocation of the search tool. Upon generating \texttt{<end\_of\_query>}, the process pauses, allowing the extraction and use of the query for retrieval. The retrieved documents are encapsulated within \texttt{<begin\_of\_documents>...<end\_of\_documents>} tags and integrated into the model's reasoning. This method ensures that retrieval is seamlessly incorporated into the reasoning process, allowing the model to continue its reasoning based on the retrieved documents without disruption.

\begin{promptbox}[System Prompt for Base Model]{lightgreen}{prompt:system}
The User asks a question, and the Assistant solves it. 
The Assistant first thinks about the reasoning process in the mind and then provides the User with the final answer. 
The output format of reasoning process and final answer are enclosed within <think> </think> and <answer> </answer> tags, respectively, i.e., "<think> reasoning process here </think><answer> final answer here </answer>". 
During the thinking process, **the Assistant can perform searching** for uncertain knowledge if necessary with the format of "<|begin\_of\_query|> search query (only list keywords, such as "keyword\_1 keyword\_2 ...")<|end\_of\_query|>". **A query must involve only a single triple**.
Then, the search system will provide the Assistant with the retrieval information with the format of "<|begin\_of\_documents|> ...search results... <|end\_of\_documents|>".
\end{promptbox}

\paratitle{Retrieve Mask-based Loss Calculation.} During the training process, the aforementioned solutions are employed to compute the RL loss, involving the reward, KL divergence, and advantages. When the model performs retrieval, the retrieved documents are integrated into the reasoning process, serving as environment observations. The model is not intended to generate these documents. To mitigate the environmental effect, we designate \texttt{<begin\_of\_documents>...<end\_of\_documents>} as special tokens and mask them during training. This prevents these external tokens from influencing the loss calculation, ensuring that the retrieved documents do not interfere with the model’s intrinsic reasoning and generation processes.

%% file: sections/exp.tex
\section{Experiment}
\label{exp}

\subsection{Datasets and Evaluation Metrics}
In training the R1-Searcher, we perform data selection from the training sets of HotpotQA and 2WikiMultiHopQA~(see \ref{tab:training_set}). We evaluate using four multi-hop datasets: HotpotQA~\cite{yang2018hotpotqa}, 2WikiMultiHopQA~\cite{ho2020constructing}, Musique~\cite{trivedi2022musique}, and Bamboogle~\cite{press2023measuring}. HotpotQA and 2WikiMultiHopQA are in-domain benchmarks since parts of their training sets are used for reinforcement learning. In contrast, Musique and Bamboogle serve as out-of-domain benchmarks to assess our model's generalization capabilities.

For evaluation metrics, following existing work~\cite{sun2025rearterretrievalaugmentedreasoningtrustworthy}, we utilize Cover Exact Match (ACC\_R) and LLM-as-Judge (ACC\_L), given the nature of open-ended multi-hop questions. Cover Exact Match assesses whether the ground truth answer is included in the predicted answer, while LLM-as-Judge uses GPT-4o-mini to evaluate the correctness of the predictions. The evaluation prompt for ACC\_L is as follows:
\begin{promptbox}{lightgreen}{prompt:judge}

Given a Question and its Golden Answer, verify whether the Predicted Answer is correct. 
The prediction is correct if it fully aligns with the meaning and key information of the Golden Answer. 
Respond with True if the prediction is correct and False otherwise.

Question: {}

Golden Answer: {}

Predicted Answer: {}
\end{promptbox}

\subsection{Baselines}
We utilize Qwen-2.5-7B-Base and Llama-3.1-8B-Instruct as the backbone models for our training. We compare R1-Searcher against the following baselines, based on GPT-4o-mini and Llama-3.1-8B-Instruct:

$\bullet$ \textbf{Naive Generation:} Direct generation of answers without retrieval.

$\bullet$ \textbf{Standard RAG:} Traditional retrieval-augmented generation systems.

$\bullet$ \textbf{Branching Methods (Branching):} SuRe~\cite{kim2024sure} and REPLUG~\cite{shi2023replug}, which execute multiple reasoning paths in parallel for a single query.

$\bullet$ \textbf{Summarization-based Methods (Summary):} LongLLMLingua~\cite{jiang2023longllmlingua}, RECOMP~\cite{xu2024recomp}, and Selective-Context~\cite{li-etal-2023-compressing}, which employ compressors to summarize retrieved documents.

$\bullet$ \textbf{Adaptive Retrieval Methods (AR):} SKR~\cite{wang2023self}, which adaptively retrieves based on the generator’s knowledge.

$\bullet$ \textbf{RAG-CoT Methods (RAG-CoT):} Self-Ask~\cite{press2022measuring}, Iter-RetGen~\cite{shao2023enhancing}, and IRCoT~\cite{trivedi2023interleaving}, integrating retrieval-augmented generation with chain-of-thought reasoning.

$\bullet$ \textbf{Test-time Scaling Methods (Test-Time):} CR-Planner~\cite{li2024can}, ReARTeR~\cite{sun2025rearterretrievalaugmentedreasoningtrustworthy}, which scale retrieval-augmented generation at test time using Monte Carlo Tree Search (MCTS).

$\bullet$ \textbf{Reasoning Models (Reasoning):} Marco-o1-Qwen-7B~\cite{zhao2024marco} and Skywork-o1-Llama-3.1-8B~\cite{skyworkopeno12024}, employing standard retrieval-augmented generation.

\subsection{Implementation Details}\label{sec:impele_details}
All baseline models adhere to the ReARTeR framework and are evaluated using FlashRAG~\cite{jin2024flashrag}. The retrieval corpus comprises the English Wikipedia as provided by KILT~\cite{kilt} in 2019, segmented into 100-word passages with appended titles, totaling 29 million passages. We employ BGE-large-en-v1.5 as the text retriever. Given the timeliness of knowledge in Bamboogle, we utilize the Google Web Search API for online webpage search tests to further evaluate our model's generalization capabilities to online search (Section~\ref{sec:online_search}).

For our R1-Searcher, the backbone model incorporates Llama-3.1-8B-Instruct or Qwen-2.5-7B-Base. The training data of the Stage-1 includes 200 medium samples from the HotpotQA training set and 150 medium samples from the 2WikiMultiHopQA training set. And the training data of Stage-2 consists of 4561 samples from HotpotQA, with 2561 medium and 2000 hard samples (Table~\ref{tab:training_set}), and 3581 samples from 2WikiMultiHopQA, also with 1087 medium and 2500 hard samples. Each data sample undergoes 16 rollouts during training, with a train batch size of 256 and a rollout batch size of 64. The learning rate is 2e-6. We utilize DeepSpeed's Zero-2~\cite{rajbhandari2020zeromemoryoptimizationstraining}, with a sampling temperature of 1.0 and a maximum retrieval count of 8. The training epoch is set to 1, with KL divergence set to 0 for Qwen-2.5-7B-Base and 1e-4 for Llama-3.1-8B-Instruct. The discount factor $\gamma$ is set to 1 in the cumulative discounted reward calculation.

\subsection{Main Results}
\begin{table*}[t]

\centering
\resizebox{1\linewidth}{!}{
\begin{tabular}{lclcccccccc}
\toprule 

\multirow{2}{*}{\textbf{Models}} &\multirow{2}{*}{\textbf{Types}} & \multirow{2}{*}{\textbf{Methods}} & \multicolumn{2}{c}{\textbf{HotpotQA}} & \multicolumn{2}{c}{\textbf{2Wiki}} & \multicolumn{2}{c}{\textbf{Bamboogle}} & \multicolumn{2}{c}{\textbf{Musique}}  \\
\cmidrule(lr){4-5} \cmidrule(lr){6-7} \cmidrule(lr){8-9} \cmidrule(lr){10-11} 
& & & $\text{ACC}_{R}$ & $\text{ACC}_{L}$ & $\text{ACC}_{R}$ & $\text{ACC}_{L}$ & $\text{ACC}_{R}$ & $\text{ACC}_{L}$ & $\text{ACC}_{R}$ & $\text{ACC}_{L}$  \\
\midrule
 \multirow{13}{*}{\textbf{GPT}} & \multirow{2}{*}{\textbf{Zero-Shot}} & Naive Generation & 0.324 & 0.404 & 0.348 & 0.346 & 0.240 & 0.280  & 0.134 & 0.170  \\
&& Standard RAG & 0.342 & 0.450& 0.344 & 0.292 & 0.272 & 0.328  & 0.172 & 0.188\\

\cmidrule{3-3}
&\multirow{2}{*}{\textbf{Branching}} & SuRe & 0.270 & 0.380 & 0.244 & 0.264 & 0.168 & 0.208  & 0.128 & 0.146  \\
 && REPLUG & 0.350 & 0.428 & 0.296 & 0.254 & 0.224 & 0.256  & 0.132 & 0.138 \\
\cmidrule{3-3}
&\multirow{3}{*}{\textbf{Summary}} & LongLLMLingua & 0.358 & 0.450 & 0.324 & 0.316 & 0.248 & 0.288  & 0.150 & 0.172  \\
 && RECOMP & 0.332 & 0.398 & 0.298 & 0.306 & 0.136 & 0.176  & 0.118 & 0.134  \\
 && Selective-Context & 0.366 & 0.442 & 0.350 & 0.290 & 0.240 & 0.288  & 0.152 & 0.172  \\
\cmidrule{3-3}
&\multirow{1}{*}{\textbf{Adaptive}} & SKR & 0.360 & 0.454 & 0.364 & 0.314 & 0.248 & 0.288  & 0.162 & 0.174  \\
\cmidrule{3-3}
&\multirow{3}{*}{\textbf{RAG-CoT}} & Self-Ask & 0.392 & 0.462& 0.336 & 0.478 & 0.336 & 0.416  & 0.260 & 0.270 \\
 && Iter-RetGen & 0.374 & 0.456 & 0.326 & 0.270 & 0.232 & 0.256  & 0.178 & 0.188\\
 && IRCoT & 0.434 & 0.308 & 0.492 & 0.114 & 0.272 & 0.184  & 0.192 & 0.214  \\
\cmidrule{3-3}
&\multirow{2}{*}{\textbf{Test-Time}} & CR-Planner  & 0.404 & 0.416 & 0.520 & 0.478 & 0.488 & 0.524 & 0.272 & 0.262 \\
&&  ReARTeR &{0.468} & {0.506} & {0.554} & {0.534} & {0.496} & {0.544}  & \textbf{0.296} & {0.302} \\
\midrule
\multirow{15}{*}{\textbf{Llama}} & \multirow{2}{*}{\textbf{Zero-Shot}} & Naive Generation  & 0.208 & 0.268 & 0.326 & 0.254 & 0.144 & 0.168 & 0.068 & 0.096  \\
 && Standard RAG  & 0.334 & 0.398 & 0.336 & 0.212 & 0.168 & 0.216 & 0.104 & 0.098\\
\cmidrule{3-3}
&\multirow{2}{*}{\textbf{Branching}} & SuRe & 0.266 & 0.346 & 0.122 & 0.262 & 0.160 & 0.192  & 0.106 & 0.144  \\
 && REPLUG & 0.290 & 0.348 & 0.334 & 0.204 & 0.168 & 0.232  & 0.078 & 0.090  \\
\cmidrule{3-3}
&\multirow{3}{*}{\textbf{Summary}} & LongLLMLingua & 0.314 & 0.382 & 0.304 & 0.294 & 0.168 & 0.216  & 0.088 & 0.100  \\
 && RECOMP & 0.318 & 0.380  & 0.324 & 0.322 & 0.104 & 0.160 & 0.112 & 0.126  \\
 && Selective-Context & 0.296 & 0.358 & 0.266 & 0.204 & 0.144 & 0.200  & 0.092 & 0.104  \\
\cmidrule{3-3}
&\multirow{1}{*}{\textbf{Adaptive}} & SKR & 0.300 & 0.372 & 0.336 & 0.212 & 0.176 & 0.208  & 0.100 & 0.112 \\
\cmidrule{3-3}
&\multirow{3}{*}{\textbf{RAG-CoT}} & Self-Ask & 0.316 & 0.408  & 0.306 & 0.322 & 0.360 & 0.432& 0.222 & 0.226  \\
 && Iter-RetGen & 0.302 & 0.362 & 0.310 & 0.224 & 0.144 & 0.176  & 0.084 & 0.084  \\
 && IRCoT & 0.210 & 0.146 & 0.338 & 0.312 & 0.120 & 0.104  & 0.060 & 0.042 \\
\cmidrule{3-3}
&\multirow{2}{*}{\textbf{Test-Time}} & CR-Planer & 0.332 & 0.350 & 0.420 & 0.350 & 0.304 & 0.336  & 0.144 & 0.098 \\
&& ReARTeR  & {0.424} & {0.434} & {0.470} & {0.364} & {0.438} & {0.484}  &{0.244} & {0.252} \\
\cmidrule{3-3}
&\multirow{2}{*}{\textbf{Reasoning}} & Marco-o1 & 0.352 & 0.348 & 0.442 & 0.184 & 0.224 & 0.200  & 0.134 & 0.104 \\
 && Skywork-o1 & 0.306 & 0.256 & 0.344 & 0.190 & 0.176 & 0.160  & 0.092 & 0.060  \\
\midrule

\textbf{Llama} & \textbf{RL} & \multirow{2}{*}{{R1-Searcher}} & 0.648 & 0.746 & 0.594 & 0.628 & 0.504 & \textbf{0.544}  & 0.254 & 0.282 \\
\textbf{Qwen}&\textbf{RL-Zero} & & \textbf{0.654} & \textbf{0.750}& \textbf{0.636} & \textbf{0.650} & \textbf{0.528} & \textbf{0.544}  & {0.282} & \textbf{0.314} \\ 

\bottomrule
\end{tabular}
}
\caption{Performance comparisons between R1-Searcher and the baselines on four multi-hop QA benchmarks. The \textbf{boldface} indicates the best performance. \textit{GPT}, \textit{Qwen}, and \textit{Llama} are the abbreviations of GPT-4o-mini, Qwen-2.5-7B-Base, and Llama-3.1-8B-Instruct, respectively.}
\label{tab:main_results} 
\end{table*}

Table\ref{tab:main_results} shows the results of R1-Searcher and the baselines on four mutil-step benchmarks. We can obtain the following observations: 

$\bullet$ \emph{Achieving Significant Performance Improvement on Multi-Hop QA.} 
ReARTeR demonstrates superior performance among existing baselines, highlighting the advantages of the test-time scaling method. However, it relies on MCTS for solution exploration, which incurs significant overhead due to increased retrieval invocations. In contrast, our proposed R1-Searcher, utilizing the same LLaMA-3.1-8B-Instruct backbone model, achieves notable performance enhancements over ReARTeR and other baselines. Specifically, our method yields improvements of 48.2\% on HotpotQA, 21.7\% on 2WikiMultiHopQA, and 4.0\% on Bamboogle according to the LLM-as-Judge metric. This indicates that our method can efficiently facilitates the model to conduct accurate retrieval invocations during the reasoning process.

$\bullet$ \emph{Supporting RL Learning from Base LLM without Cold Start.} Furthermore, we also conduct RL learning from scratch using a powerful base model, such as Qwen-2.5-7B-Base. Surprisingly, we can achieve better results and obtain the best performance on most in-domain and out-of-domain datasets, even surpassing the closed-source LLM such as GPT-4o-mini. These results demonstrate the effectiveness of our two-stage RL method in guiding the LLMs' learning process.

$\bullet$ \emph{Maintaining Generalization Ability.} We employ only 8148 samples from the training sets of HotpotQA and 2WikiMultiHopQA for RL training. The model not only excels on these in-domain datasets but also demonstrates strong generalization by performing well on the out-of-domain datasets, such as Musique and Bamboogle. This suggests that the model effectively learns retrieval and integrates it with reasoning through exploration during RL training, enabling robust performance on new test datasets requiring retrieval. Furthermore, it can also seamlessly generalizes to online search, as detailed in Section~\ref{sec:online_search}.

%% file: sections/research_question.tex
\section{Further Analysis}

In this section, we present a detailed discussion of several key aspects that should be considered during the training process.

\subsection{Basic Training Methods}
\begin{figure}[ht]
\centering
    \centering
    \hspace{-2mm}{
    \includegraphics[width=1.0\linewidth]{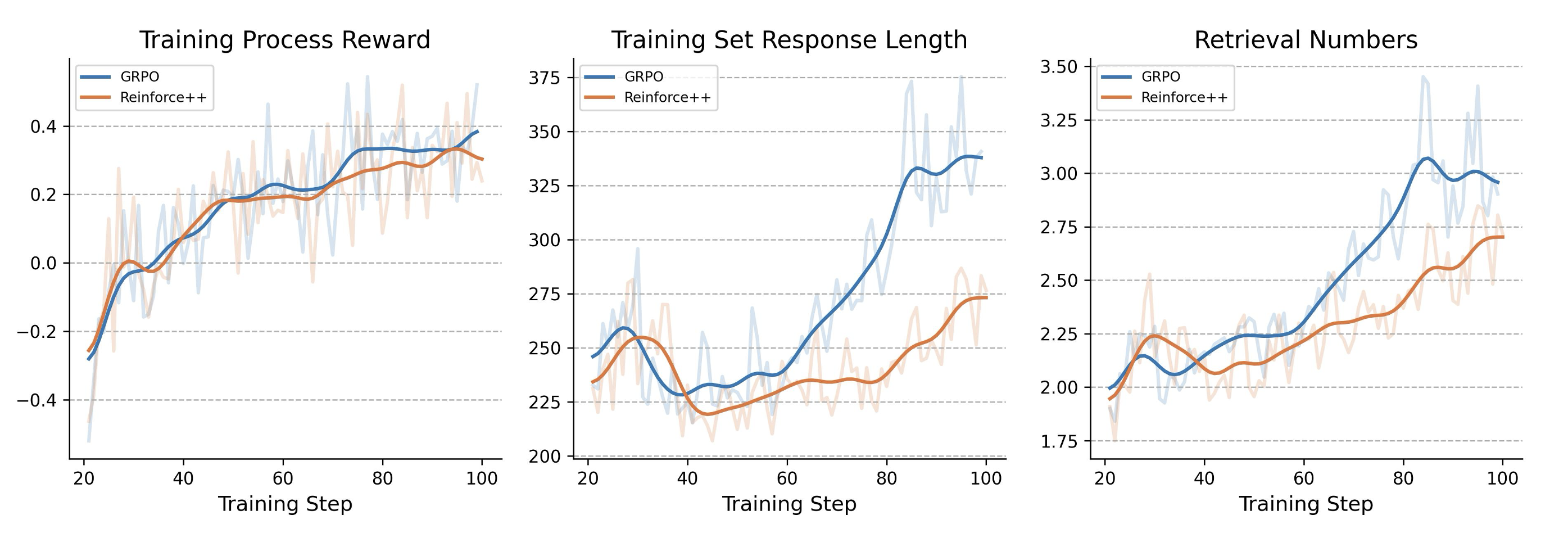}}
\caption{The log of reward, response length, and retrieval numbers for Llama-3.1-8B-Instruct comparison between using GRPO and Reinforce++.
}
    \label{fig:4.1.1}
\end{figure}

\paratitle{GRPO or Reinforce++.}
As two representative RL algorithms that do not require a critic model, we compare the differences between GRPO~\cite{shao2024deepseekmathpushinglimitsmathematical} and Reinforce++ on our RAG tasks.
We perform two-stage training on Llama-3.1-8B-Instruct, setting the KL divergence to $1e-4$ and utilizing HotpotQA and 2Wiki as the training datasets.
As shown in Figure~\ref{fig:4.1.1}, 
although there are no significant differences in rewards between the two algorithms during training, GRPO demonstrates a clear advantage in both the length of generated text and the frequency of retrievals. 
The generation of longer text may widen the reasoning scope, and the increased frequency of retrievals could potentially improve the accuracy in responding to queries where the model itself has uncertainty.
Moreover, it also demonstrates better performance on the out-of-domain dataset (\ie{Bamboogle}), suggesting that GRPO may possess superior generalization capabilities.
However, Reinforce++ exhibits superior performance on the in-domain test set~(\ie HotpotQA and 2Wiki), which seemingly indicates a higher learning efficiency towards in-domain data.

\begin{table*}[ht]
\centering
\small
\begin{tabular}{lcccccccccc}
\toprule
\multicolumn{1}{c}{\multirow{2.5}{*}{\textbf{Method}}} & \multicolumn{3}{c}{\textbf{HotpotQA}} & \multicolumn{3}{c}{\textbf{2Wiki}} & \multicolumn{3}{c}{\textbf{Bamboogle}} & \multicolumn{1}{c}{\multirow{2.5}{*}{\textbf{Avg (CEM)}}}   \\
\cmidrule(r){2-4} \cmidrule(r){5-7} \cmidrule(r){8-10} 
\multicolumn{1}{c}{} & $\text{EM}$ & $\text{CEM}$ & $\text{F1}$ & $\text{EM}$ & $\text{CEM}$ & $\text{F1}$ & $\text{EM}$ & $\text{CEM}$ & $\text{F1}$\\
\midrule
GRPO & 53.0 &  60.5 &68.6 &58.0 & 60.5 & 63.0 &  48.0 & 56.0 &  60.5 & 59.0\\
Reinforce++ & 58.4 & 64.8 &  70.6 & 57.5 &  61.5 &  62.9 & 44.0 & 50.4 & 57.1&58.9 \\
\bottomrule
\end{tabular}
\caption{
Performance comparison of Llama-3.1-8B-Instruct trained using GRPO and Reinforce++ on three multi-hop QA benchmarks.}
\label{tab:4.1.1}
\end{table*}

\paratitle{RL or SFT.}
In this part, we aim to understand the enhancement effects of SFT and RL through comparison.
We conduct RL training according to the same settings in Section ~\ref{sec:impele_details}. For the SFT data, we select Qwen-2.5-7B-instruct and conduct rollouts from the training sets of HotpotQA and 2Wiki, obtaining 4768 pieces of data with good reasoning paths. Among them, 4268 pieces of data undergo retrieval, and the training epoch is set to 3. The results are shown in Table~\ref{tab:4.1.2}. We can see that RL outperforms SFT in both in-domain and out-of-domain test sets, indicating superior retrieval capability and generalization across varying datasets.
After inspecting the outputs of models trained with both methods (see Section~\ref{case_sft_rl}), we find that although SFT assists the model in generating retrieval queries, the timing and relevance of these queries are inferior to those produced by RL training. Specifically, SFT tends to rely on the model's internal knowledge, which can often be erroneous or misleading.
This indicates that RL may be more effective in enhancing the model's retrieval skills.

\begin{table*}[ht]
\centering
\small
\begin{tabular}{lccccccccccc}
\toprule
\multicolumn{1}{c}{\multirow{2.5}{*}{\textbf{Method}}} & \multicolumn{3}{c}{\textbf{HotpotQA}} & \multicolumn{3}{c}{\textbf{2Wiki}} & \multicolumn{3}{c}{\textbf{Bamboogle}}  & \multicolumn{1}{c}{\multirow{2.5}{*}{\textbf{\makecell{Avg\\(CEM)}}}} \\
\cmidrule(r){2-4} \cmidrule(r){5-7} \cmidrule(r){8-10} 
\multicolumn{1}{c}{} & $\text{EM}$ & $\text{CEM}$ & $\text{F1}$ & $\text{EM}$ & $\text{CEM}$ & $\text{F1}$ & $\text{EM}$ & $\text{CEM}$ & $\text{F1}$\\
\midrule
Qwen-Base-RL & 58.0 & 65.4 & 71.9 & 55.4 & 63.6 & 63.7 & 45.6 & 52.8 & 57.7 & 60.6\\
Qwen-Base-SFT & 37.0 &49.5 &  51.3 & 42.5 & 54.5 &  51.3 & 40.8 &  46.4 & 51.0 & 50.1 \\
Llama-Instruct-RL & 58.4 & 64.8 &  70.6 & 55.0 &  59.4 &  61.2 & 44.0 & 50.4 & 57.1 & 58.2\\
Llama-Instruct-SFT & 36.0 & 47.0 & 50.4 & 38.0 & 51.0 & 48.3 &39.4 & 46.6 & 48.2 & 48.2\\
\bottomrule
\end{tabular}
\caption{Performance comparison of Qwen-2.5-7B-Base and Llama-3.1-8B-Instruct trained using RL and SFT on three multi-hop QA benchmarks. \textit{Qwen-Base} and \textit{Llama-Instruct} are the abbreviations of  Qwen-2.5-7B-Base and Llama-3.1-8B-Instruct, respectively.}
\label{tab:4.1.2}
\end{table*}

\subsection{Reward Design}
\begin{figure}[ht]
\centering
    \centering
    \hspace{-2mm}{
    \includegraphics[width=1.0\linewidth]{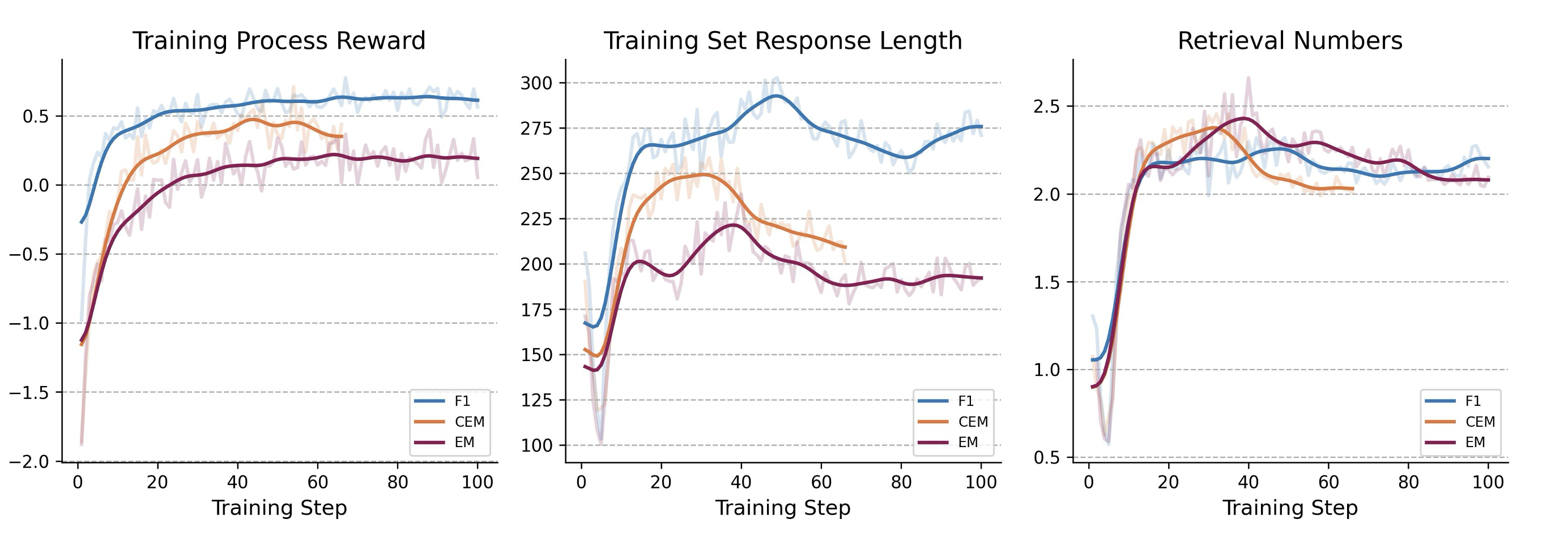}}
\caption{The log of reward, response length, and retrieval numbers for the Qwen-2.5-7B-Base model utilizing different metrics for outcome-supervised reward calculation.}
    \label{fig:4.2.1}
\end{figure}

\paratitle{Answer Reward.}
Here, we investigate the impact of various answer rewards on RL training. We specifically compare the performance of using Exact Match (EM), Cover Exact Match (CEM), and F1 score as answer rewards. The F1 score is used directly as its own reward, while the rewards for EM and CEM are defined as follows:
\begin{equation}
    R_{answer} =
\begin{cases}
\text{1}, & \text{if EM/CEM is True} \\
\text{-1}, & \text{if EM/CEM is False}
\end{cases}
\end{equation}
The training log and final results are presented in Figure~\ref{fig:4.2.1} and Table~\ref{tab:4.2.1}. Firstly, the F1-based answer reward yields longer response lengths and superior final results compared to CEM and EM-based rewards. Notably, it achieves up to a 52.6\% average performance improvement over the EM-based reward. Secondly, the EM-based reward results in shorter response lengths during training and poorer performance during testing compared to CEM or F1-based reward. This may be due to EM's strictness, making it unsuitable for open-ended question generation scenarios. Overall, F1 provides a more balanced measure of answer accuracy, serving as a more effective outcome-based reward in this scenario.

\begin{table*}[ht]
\centering
\small
\begin{tabular}{lcccccccccc}
\toprule
\multicolumn{1}{c}{\multirow{2.5}{*}{\textbf{Method}}} & \multicolumn{3}{c}{\textbf{HotpotQA}} & \multicolumn{3}{c}{\textbf{2Wiki}} & \multicolumn{3}{c}{\textbf{Bamboogle}} & \multicolumn{1}{c}{\multirow{2.5}{*}{\textbf{Avg (CEM)}}}  \\
\cmidrule(r){2-4} \cmidrule(r){5-7} \cmidrule(r){8-10} 
\multicolumn{1}{c}{} & $\text{EM}$ & $\text{CEM}$ & $\text{F1}$ & $\text{EM}$ & $\text{CEM}$ & $\text{F1}$ & $\text{EM}$ & $\text{CEM}$ & $\text{F1}$\\
\midrule
EM & 55.0 & 62.0 & 69.3 & 29.0 & 29.0 & 30.0 &  24.8 & 28.0 & 33.2&39.7 \\
CEM & 53.4 & 65.0 &68.8 & 51.8 & 59.2 & 61.7 &46.4 & 54.4 & 59.0 & 59.5\\
F1 & 58.0 & 65.4 & 71.9 & 55.4 & 63.6 & 63.7 & 45.6 & 52.8 & 57.7 & 60.6\\
\bottomrule
\end{tabular}
\caption{Performance comparison of the Qwen-2.5-7B-Base model utilizing different metrics for outcome-supervised reward calculation on three mutil-hop QA benchmarks.}
\label{tab:4.2.1}
\end{table*}




\paratitle{Format Reward.}
During training, we impose strict constraints on the format reward (see Section~\ref{format_reward}). These constraints are iteratively refined to address instances of reward hacking and the generation of unreasonable solutions. The primary issues observed include:
\begin{enumerate}
\item The model produces \texttt{<begin\_of\_documents>...<end\_of\_documents>} without generating \texttt{<begin\_of\_query>...<end\_of\_query>}, effectively creating ``external documents'' independently.

\item When training with the Base model and setting KL to 0, the model occasionally generates nonsensical output in later training phases, failing to adhere to specified formats.


\item With the Llama model, omitting the Stage-1 training causes the model to bypass retrieval entirely, directly answering questions without engaging in the retrieval process.

\item Using CEM as the supervisory signal, the model often produces lengthy responses containing extraneous information, though the correct answer is included.
\end{enumerate}
Through our designed format rewards, we can train the model more stably in the RL training process, avoiding abnormal outputs and reward hacking.

\subsection{Training Data}
\begin{figure}[ht]
\centering
    \centering
    \hspace{-2mm}{
    \includegraphics[width=1.0\linewidth]{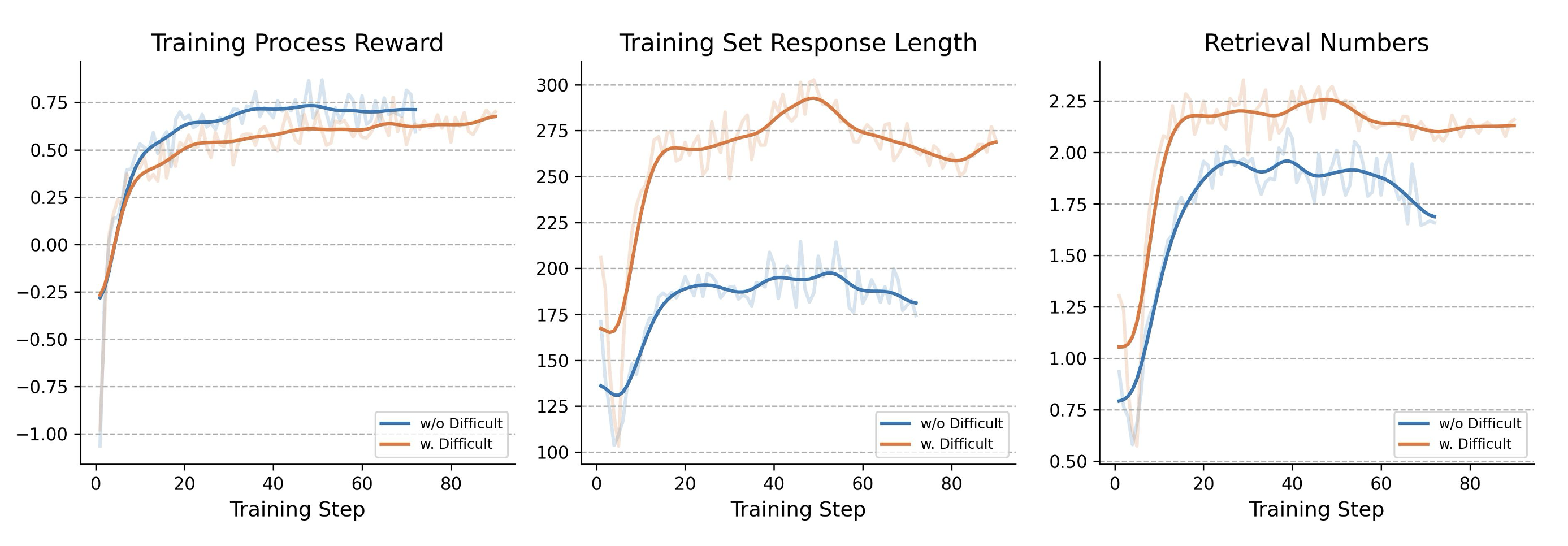}}
\caption{The log of reward, response length, and retrieval numbers for the Qwen-2.5-7B-Base model, trained on datasets of varying difficulty levels.}
    \label{fig:4.3.1}
\end{figure}

\paratitle{Difficulty Distribution.} In this study, we examine the effect of data difficulty on training by constructing two distinct datasets. The first dataset, used for primary training, is labeled \textit{w. Difficult} (Table~\ref{tab:training_set}). The second dataset, \textit{w/o Difficult}, substitutes questions requiring more than 20 rollouts with those requiring 10 to 20 rollouts. Both datasets are trained under identical configurations.
As shown in Figure~\ref{fig:4.3.1}, training with the \textit{w/o Difficult} dataset results in shorter generation lengths and fewer retrievals compared to the \textit{w. Difficult} dataset. This suggests that more challenging problems prompt the model to perform additional retrievals to answer questions. Furthermore, Table~\ref{tab:4.3.1} indicates that models trained on the \textit{w. Difficult} dataset achieves superior performance on the evaluation dataset compared to those trained on the \textit{w/o Difficult} dataset~(achieving 3.4\% average CEM performance improvements on three datasets). This underscores the importance of data difficulty distribution for model performance in RL, as more challenging questions enhance the model's reasoning capabilities.

\begin{table*}[ht]
\centering
\small
\begin{tabular}{lcccccccccc}
\toprule
\multicolumn{1}{c}{\multirow{2.5}{*}{\textbf{Method}}} & \multicolumn{3}{c}{\textbf{HotpotQA}} & \multicolumn{3}{c}{\textbf{2Wiki}} & \multicolumn{3}{c}{\textbf{Bamboogle}} & \multicolumn{1}{c}{\multirow{2.5}{*}{\textbf{Avg (CEM)}}}  \\
\cmidrule(r){2-4} \cmidrule(r){5-7} \cmidrule(r){8-10} 
\multicolumn{1}{c}{} & $\text{EM}$ & $\text{CEM}$ & $\text{F1}$ & $\text{EM}$ & $\text{CEM}$ & $\text{F1}$ & $\text{EM}$ & $\text{CEM}$ & $\text{F1}$\\
\midrule
w/o Difficult & 54.8 & 61.8 & 69.3 & 55.4 & 63.6 & 63.7 &  44.8 &51.2 & 56.9 &58.8 \\
w. Difficult & 58.0 & 65.4 & 71.9 & 54.8 & 64.2 & 63.8 & 45.6 & 52.8 & 57.7& 60.8 \\
\bottomrule
\end{tabular}
\caption{Performance comparison of the Qwen-2.5-7B-Base model trained on datasets of different difficulty levels on three mutil-hop QA benchmarks.}
\label{tab:4.3.1}
\end{table*}



\begin{figure}[ht]
\centering
    \centering
    \hspace{-2mm}{
    \includegraphics[width=1.0\linewidth]{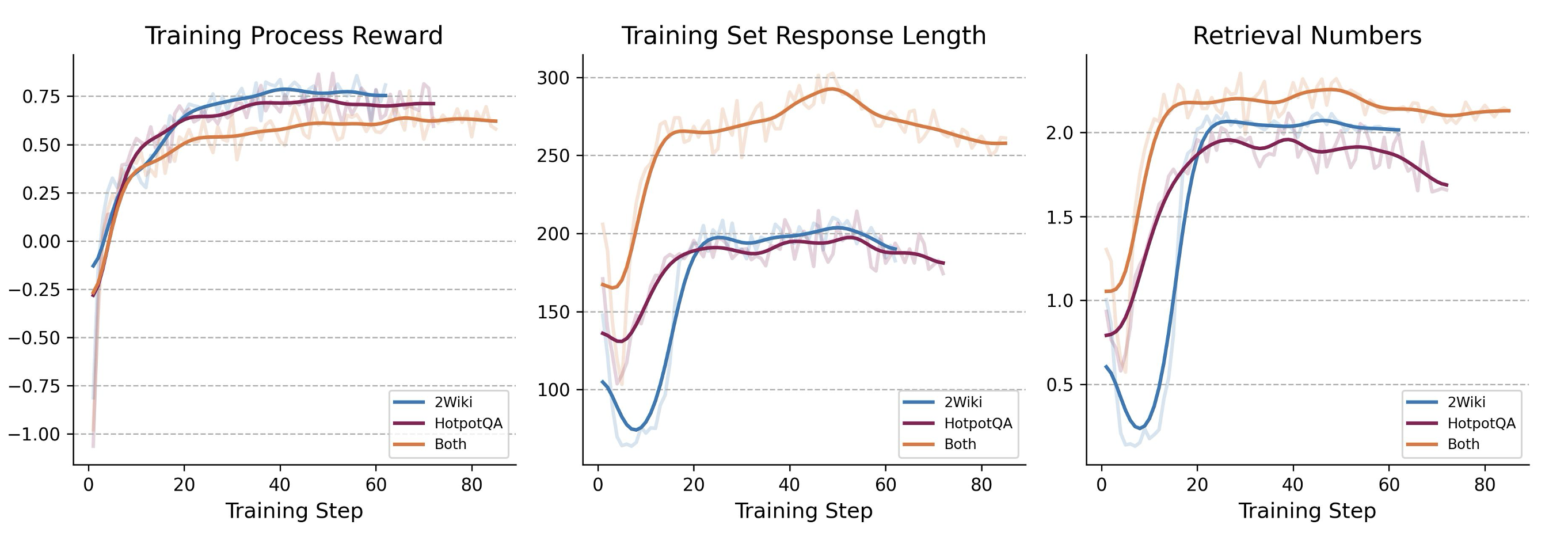}}
\caption{The log of reward, response length, and retrieval numbers for the Qwen-2.5-7B-Base model trained on different datasets.}
    \label{fig:4.3.2}
\end{figure}

\paratitle{Data Diversity.}
We investigate the effect of data diversity during the RL training process.
Specifically, we compare the performance of using a combination of the HotpotQA and 2Wiki datasets, as well as each dataset individually. The training log and final results are presented in Figure~\ref{fig:4.3.2} and Table~\ref{tab:4.3.2}, respectively.
We can find that models trained on the mixed dataset show an increase in the number of retrievals and the length of generated responses compared to those trained on either dataset alone, achieving higher scores on the test set, with improvements of up to 10.9\% in average CEM performance.
Additionally, models trained solely on the 2Wiki dataset demonstrate superior training rewards but inferior average performance across three datasets compared to those trained on the HotpotQA dataset. This may be attributed to the relatively low diversity within the 2Wiki dataset, potentially leading to overfitting during RL training. These findings demonstrate that the diversity of training datasets significantly affects both training efficacy and generalizability, underscoring the importance of data diversity.

\begin{table*}[ht]
\centering
\small
\begin{tabular}{lcccccccccc}
\toprule
\multicolumn{1}{c}{\multirow{2.5}{*}{\textbf{Method}}} & \multicolumn{3}{c}{\textbf{HotpotQA}} & \multicolumn{3}{c}{\textbf{2Wiki}} & \multicolumn{3}{c}{\textbf{Bamboogle}} & \multicolumn{1}{c}{\multirow{2.5}{*}{\textbf{Avg (CEM)}}}  \\
\cmidrule(r){2-4} \cmidrule(r){5-7} \cmidrule(r){8-10} 
\multicolumn{1}{c}{} & $\text{EM}$ & $\text{CEM}$ & $\text{F1}$ & $\text{EM}$ & $\text{CEM}$ & $\text{F1}$ & $\text{EM}$ & $\text{CEM}$ & $\text{F1}$\\
\midrule
HotpotQA & 53.8 & 59.2 & 67.2 & 46.7 & 54.3 & 54.7 & 44.0 & 50.4 &  55.1 &54.6 \\
2Wiki & 46.0 &50.5 & 58.7 & 45.0 & 47.5 & 48.2 & 31.2 & 32.8 & 39.4 &43.6 \\
Mixture & 58.0 & 65.4 & 71.9 & 55.4 & 63.6 & 63.7 & 45.6 & 52.8 & 57.7 &  60.6\\
\bottomrule
\end{tabular}
\caption{Performance comparison of the Qwen-2.5-7B-Base model trained on different datasets.
}
\label{tab:4.3.2}
\end{table*}




\subsection{Generalization to Online Search}\label{sec:online_search}

\begin{figure}[ht]
\centering
    \centering
    \includegraphics[width=0.9\linewidth]{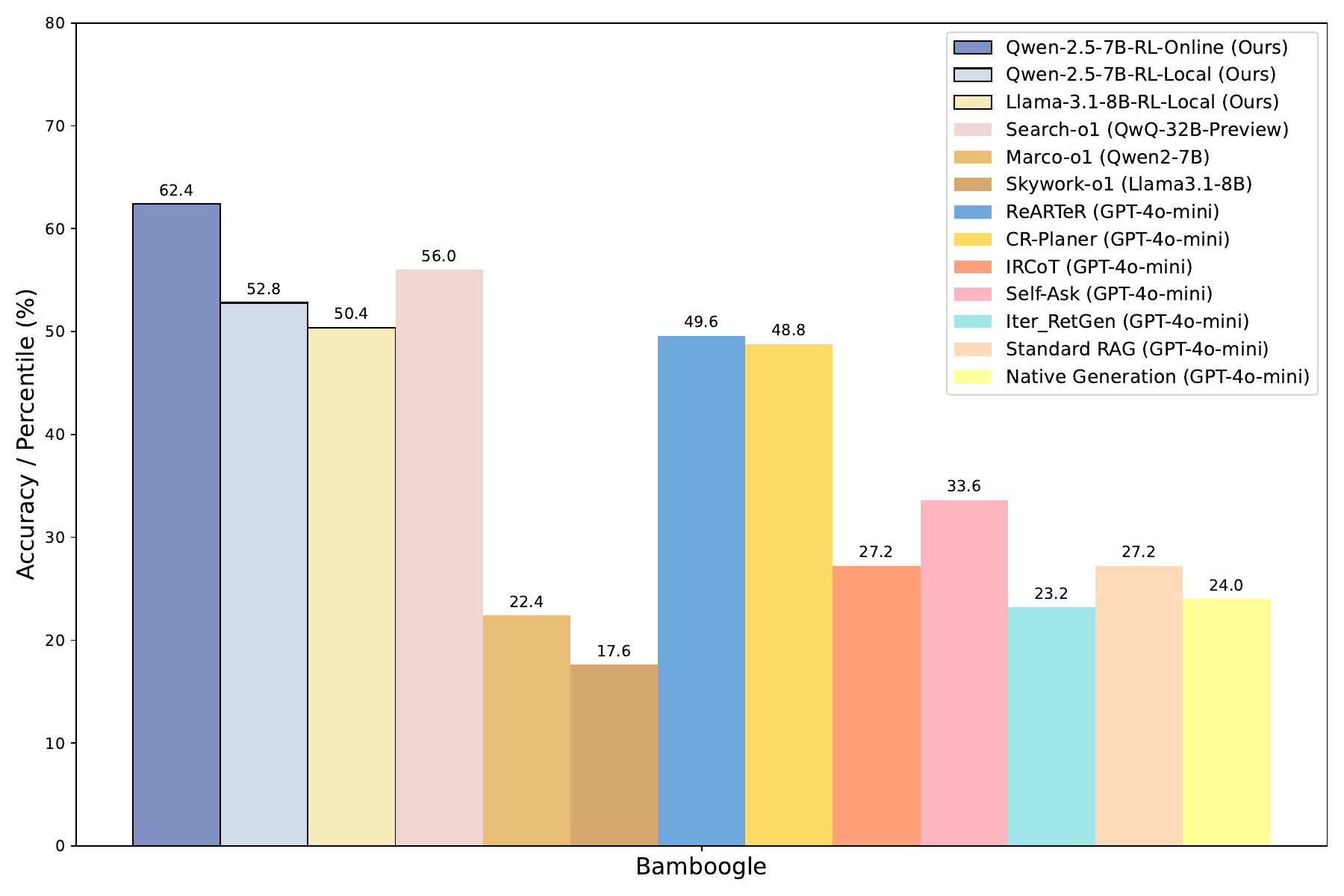}
\caption{Preference comparison of our models that utilize local search and online search and the baselines on the Bamboogle dataset. Search-o1 utilizes online search, and all other baselines employ local search. The metric is CEM (Cover Exact Match). 
}
    \label{fig:bamboogle-web}
\end{figure}
In consideration of training speed and cost, we implement a local dense embedding-based retrieval system using Wikipedia as the external retrieval environment. To evaluate the model's generalization capabilities beyond this knowledge scope, we test its performance on the newly proposed Bamboogle task using an online search scenario, which is not encountered during the RL training. Specifically, when retrieval is required, we employ the Google API to perform online searches, obtaining relevant web pages. Due to the extensive length of these pages, we first summarize them using GPT-4o-mini before integrating the summaries into the reasoning model. As illustrated in Figure~\ref{fig:bamboogle-web}, our model demonstrates an 18.2\% improvement over the local retrieval system with the same Qwen-2.5-7B-Base backbone. Additionally, it achieves an 11.4\% performance increase compared to Search-o1, which utilizes a larger 32B backbone model with the same online search. This indicates our model's adaptability to online search scenarios and shows that R1-Searcher enables the model to retrieve information during reasoning rather than merely memorizing response formats.


%% file: sections/case_study.tex
\section{Case Study}

\subsection {RL v.s. SFT}\label{case_sft_rl}

Compared to SFT, RL is better at invoking external retrieval for querying when it doesn't know how to answer by itself. However, SFT is prone to being influenced by internal knowledge and directly uses the incorrect internal knowledge to answer, ultimately resulting in wrong answers to the questions. We have presented a detailed comparison as shown in Table~\ref{sft_vs_rl_1}.

\subsection {Qwen v.s. Llama}

As shown in Table~\ref{qwen_vs_llama}, after RL, the Qwen-2.5-7B-Base model more effectively decomposes complex questions than the Llama-3.1-8B-Instruct model. It refines reasoning by breaking down multi-step problems into subproblems and sequentially retrieving relevant information, thereby increasing the likelihood of retrieving pertinent documents and enhancing accuracy.

\begin{table}[h]
\centering
\fontsize{9}{10}\selectfont
\begin{tabular}{p{13cm}}
\toprule
Question: \\
When did the last king from Britain's House of Hanover die? \\
Answer: \\
20 June 1837 \\
\midrule
Response (RL) \textcolor{green}{\textit{\textbf{ (True)}}}:

To solve this question, I need to find out when the last king from Britain's House of Hanover died. 

Step 1: Identify the name of the last king from the House of Hanover.
Step 2: Determine the date of the last king's death.

**Step 1: Identify the name of the last king from the House of Hanover.**

I'll use a search query to find out:

<|begin\_of\_query|> last king from Britain's House of Hanover<|end\_of\_query|>

<|begin\_of\_documents|>

\ldots \\
<|end\_of\_documents|>

From the search results, I can see that the last British monarch who reigned in Hanover was William IV.

**Step 2: Determine the date of the last king's death.**

I'll use a search query to find out when William IV died:

<|begin\_of\_query|> William IV of Hanover death date<|end\_of\_query|>

<|begin\_of\_documents|>

\ldots \\
<|end\_of\_documents|>

From the search results, I can see that William IV died on 20 June 1837.

Therefore, the last king from Britain's House of Hanover died on 20 June 1837.</think>

 <answer> 20 June 1837 </answer>\\
\midrule
Response (SFT) \textcolor{red}{\textit{\textbf{ (False)}}}: 

To answer this question, I need to recall the British monarchs who belonged to the House of Hanover and identify the last one. The House of Hanover includes kings like George I, George II, and George III. I need to find out when George III, the last Hanoverian king, passed away.

George III ruled from 1760 to 1820. After his death, the throne passed to his son, George IV, who was from the House of Hanover's successor, the House of Saxe-Coburg and Gotha.

I will verify the date of George III's death to provide the correct answer.
<|begin\_of\_query|> George III death date<|end\_of\_query|>

<|begin\_of\_documents|>

\ldots \\
<|end\_of\_documents|>

George III died on 29 January 1820. This confirms that he was the last king from Britain's House of Hanover, and there was no Hanoverian king after him.

The answer is 29 January 1820.
</think> <answer> 29 January 1820 </answer>\\

\midrule
\end{tabular}
\caption{Case study from the Bamboogle dataset comparing Qwen-2.5-7B-Instruct models trained with RL and SFT.}
\label{sft_vs_rl_1}
\end{table}

\begin{table}[ht]
\centering
\fontsize{9}{10}\selectfont
\begin{tabular}{p{13cm}}
\toprule
Question: \\
Who was president of the United States in the year that Citibank was founded? \\
Answer: \\
james madison\\
\midrule
Response (Qwen-2.5-7B-Base) \textcolor{green}{\textit{\textbf{ (True)}}}:

To solve this question, I need to find out who the president of the United States was in the year that Citibank was founded and then identify that president.

Step 1: Determine the year Citibank was founded.
I'll use a search query to find out:

<|begin\_of\_query|> Citibank founding year<|end\_of\_query|>

<|begin\_of\_documents|>

\ldots \\
<|end\_of\_documents|>

From the search results, I can see that Citibank was founded in 1812.

Step 2: Identify the president of the United States in 1812.
I'll use a search query to find out:

<|begin\_of\_query|> president of the United States in 1812<|end\_of\_query|>

<|begin\_of\_documents|>

\ldots \\
<|end\_of\_documents|>

The search results mention that James Madison was the President of the United States in 1812.

Therefore, the president of the United States in the year that Citibank was founded was James Madison.</think>

 <answer> James Madison </answer>\\

\midrule

Response (Llama-3.1-8B-Instruct) \textcolor{green}{\textit{\textbf{ (True)}}}:

Given the information, I don't have specific knowledge about the president of the United States in the year Citibank was founded. I will perform a search to find the required information. The query to search for is "<|begin\_of\_query|> president of united states citibank founded<|end\_of\_query|>

<|begin\_of\_documents|>

\ldots \\
<|end\_of\_documents|>

Based on the search results, I found that Citibank was founded in 1812, and I couldn't find specific information about the president of the United States in that year. I will perform another search to find the required information. The query to search for is "<|begin\_of\_query|> president united states 1812<|end\_of\_query|>

<|begin\_of\_documents|>

\ldots \\
<|end\_of\_documents|>

Based on the search results, I found that the president of the United States in 1812 was James Madison.</think>

<answer>James Madison</answer>\\

\bottomrule
\end{tabular}
\caption{Case study from the Bamboogle dataset comparing Qwen-2.5-7B-Instruct and Llama-3.1-8B-Instruct models trained with RL.}
\label{qwen_vs_llama}
\end{table}

%% file: sections/conclusion.tex
\section{Conclusion}

In this paper, we propose R1-Searcher, a novel framework integrating RAG with RL. This framework employs a two-stage outcome-supervised RL approach, which enables the model to learn to invoke an external search engine to acquire relevant knowledge during the reasoning process through a meticulously designed reward mechanism.
The proposed method relies solely on RL, allowing the model to learn autonomously through exploration without requiring any instruction fine-tuning for cold start. It demonstrates the ability to generalize from in-domain training datasets to out-of-domain test datasets while seamlessly switching to online search to obtain up-to-date information. Moreover, R1-Searcher is applicable to both base models and instruction-tuned models.
Extensive experiments conducted on multiple datasets show that R1-Searcher outperforms traditional RAG methods and other reasoning approaches. Additionally, we analyze the training process from various aspects, including training methods, data, and reward designing.

%% file: sections/future_work.tex
\section{Future Work}
\label{future_work}
In future work, we aim to refine our training methodology in two key areas. First, we will explore more sophisticated data curricula, as we have observed that the distribution and difficulty of training data significantly influence the learning process. So far, we have only employed simple data mixing, and a more structured approach may further enhance performance. Second, we plan to scale up our model beyond the current 7B configuration, investigating larger models (e.g., 32B) to better assess the effectiveness of our approach.